# Minimal Support Vector Machine

**Shuai Zheng and Chris Ding**
University of Texas at Arlington
zhengs123@gmail.com, chqding@uta.edu

## Abstract

Support Vector Machine (SVM) is an efficient classification approach, which finds a hyperplane to separate data from different classes. This hyperplane is determined by support vectors. In existing SVM formulations, the objective function uses L2 norm or L1 norm on slack variables. The number of support vectors is a measure of generalization errors. In this work, we propose a *Minimal SVM*, which uses L0.5 norm on slack variables. The result model further reduces the number of support vectors and increases the classification performance.

## 1 Introduction

Support Vector Machine (SVM) is an efficient classification approach, which finds a hyperplane to separate data from different classes. SVM has been widely used in object classification, face recognition, text categorization and so on. In most of these cases, SVM generalization performance either matches or is significantly better than that of competing methods [1].

Suppose we have $n$ training samples from two classes $\{\mathbf{x}_i, y_i\}$, $i = 1, ..., n$, label indicator $y_i \in \{-1, 1\}$, $\mathbf{x}_i \in \mathbb{R}^{k \times 1}$, where $k$ is data dimension. In linear separable case, suppose the hyperplane which separates the two classes is $\mathbf{w}^T \mathbf{x} + b = 0$, where $\mathbf{w} \in \mathbb{R}^{k \times 1}$ is normal to the hyperplane, $\mathbf{w}^T$ is the transpose of vector $\mathbf{w}$. Let $d_+$ ($d_-$) be the shortest distance from the separating hyperplane to the closest positive (negative) example. Define the margin of a separating hyperplane to be $d_+ + d_-$. Support Vector Machine finds such a separating hyperplane with the largest margin and all the training data satisfy the following constraints:

$$\mathbf{w}^T \mathbf{x}_i + b \geq +1 \text{ for } y_i = +1, \qquad (1)$$

$$\mathbf{w}^T \mathbf{x}_i + b \leq -1 \text{ for } y_i = -1. \qquad (2)$$

Combine the two equations into one:

$$y_i(\mathbf{w}^T \mathbf{x}_i + b) - 1 \geq 0 \ \ \forall i. \qquad (3)$$

Let the distance from origin of coordinate to the hyperplane $\mathbf{w}^T \mathbf{x} + b = 0$ be $d_0$, and let $d_0 \mathbf{w}/\|\mathbf{w}\|$ be the point on the hyperplane that is closest to the origin, $\mathbf{w}/\|\mathbf{w}\|$ is a unit vector that gives the direction perpendicular to the hyperplane. Since this point is on the hyperlane, we have $\mathbf{w}^T [d_0 \mathbf{w}/\|\mathbf{w}\|] + b = 0$, thus $d_0 = |b|/\|\mathbf{w}\|$. Similarly, distance from origin to hyperplane $\mathbf{w}^T \mathbf{x} + b = -1$ is $|b + 1|/\|\mathbf{w}\|$; distance from origin to hyperplane $\mathbf{w}^T \mathbf{x} + b = +1$ is $|b - 1|/\|\mathbf{w}\|$. Hence, $d_+ = d_- = 1/\|\mathbf{w}\|$, and the margin is $2/\|\mathbf{w}\|$. Thus, for linear separable case, SVM objective is given as:

$$\min \frac{1}{2}\|\mathbf{w}\|^2, \qquad (4)$$
$$\text{s.t. } y_i(\mathbf{w}^T \mathbf{x}_i + b) - 1 \geq 0 \ \ \forall i.$$

This can be solved using constrained optimization [1]. In testing, given a test data $\mathbf{x}$, we determine the class labels using $sign(\mathbf{w}^T \mathbf{x} + b)$.

When SVM is applied to non-separable data, non-negative slack variables $\xi_i$, $i = 1, ..., n$ are introduced to the constraints Eq.(1) and Eq.(2):

$$\mathbf{w}^T \mathbf{x}_i + b \geq +1 - \xi_i \text{ for } y_i = +1, \tag{5}$$

$$\mathbf{w}^T \mathbf{x}_i + b \leq -1 + \xi_i \text{ for } y_i = -1, \tag{6}$$

$$\xi_i \geq 0, \quad \forall i. \tag{7}$$

Slack variables $\xi_i$ measures training error. To minimize training errors and integrate slack variables into objective function, the non-separable SVM is given as:

$$\min \frac{1}{2} \|\mathbf{w}\|^2 + C \sum_i \xi_i, \tag{8}$$

$$\text{s.t. } y_i(\mathbf{w}^T \mathbf{x}_i + b) \geq 1 - \xi_i$$

$$\xi_i \geq 0, \quad \forall i,$$

where $C$ is a parameter that controls the weight of penalty to errors. Those training data that satisfy $y_i(\mathbf{w}^T \mathbf{x}_i + b) = 1 - \xi_i$, with $\xi_i \geq 0$, are called support vectors. We say that the constraints of support vectors are *active*. Support vectors decides the direction of the hyperplane.

Nonlinear SVM is a generalized version of linear SVM. Suppose we have a mapping function that maps the data to some other Eculidean space $\mathcal{H}$, $\Phi : \mathbb{R}^{k \times 1} \to \mathcal{H}$. A kernel function using this mapping is $K(\mathbf{x}_i, \mathbf{x}_j) = \Phi(\mathbf{x}_i) \cdot \Phi(\mathbf{x}_j)$. Both in the training and testing process, we would only use the kernel function $K$ and there is no need to know explicitly what $\Phi$ is.

Number of support vectors is a measure of generalization errors. Reducing number of support vectors can improve model prediction capability and classification accuracy can be improved. From the objective of Eq.(8), we can see that one way to reduce number of support vectors is to increase parameter $C$. However, we found that number of support vectors in Eq.(8) is not sensitive to $C$. In this work, we propose a Minimal SVM, which uses $L_{0.5}$ norm on slack variables. In Minimal SVM, number of support vectors is sensitive to $C$. On 7 binary classification tasks from 4 datasets, Minimal SVM further reduces the number of support vectors and increases the classification accuracy.

## 2 Motivation

In this section, we use a toy data set to show that number of support vectors in Eq.(8) is not sensitive to $C$. The toy data contains 100 2-dimensional random points from two classes, with 50 points in each class. Data points of each class are randomly generated by a normal distribution function. The two classes are non-separable.

As we discussed in introduction, the hyperplane direction of SVM is determined by $\mathbf{w}$ and $b$. The width of margin is $2/\|\mathbf{w}\|$. Parameter $C$ controls the weights of non-separable data errors. Figure 1a, 1c, 1e show the results using objective Eq.(8) when $C = 1$, 50, and 100. The solid black line is line $\mathbf{w}^T \mathbf{x} + b = 0$. The two dash black lines are $\mathbf{w}^T \mathbf{x} + b = -1$ and $\mathbf{w}^T \mathbf{x} + b = 1$. Two classes are denoted in blue circle and red triangle. Support vectors are those points with black squares.

From Figure 1a, 1c, 1e, we can see that the number of support vectors can be further reduced and the number of support vectors is 15 when $C = 1$ and 14 when $C = 50, C = 100$. The width of margin is decreased when $C$ increases. $2/\|\mathbf{w}\|$ is 1.7752 when $C = 1$, 1.6305 when $C = 50$, and 1.6309 when $C = 100$.

## 3 Minimal Support Vector Machine

$L_p$ norm is a generalized version of $L_1$ and $L_2$ norm. When $0 \leq p \leq 1$, $L_p$ norm introduces sparsity and has been used for feature selection [7]. In this chapter, we propose to solve the following Minimal Support Vector Machine (Minimal SVM) objective:

$$\min \frac{1}{2} \|\mathbf{w}\|^2 + C \sum_i \xi_i^p, \tag{9}$$

$$\text{s.t. } y_i(\mathbf{w}^T \mathbf{x}_i + b) \geq 1 - \xi_i$$

$$\xi_i \geq 0 \quad \forall i.$$

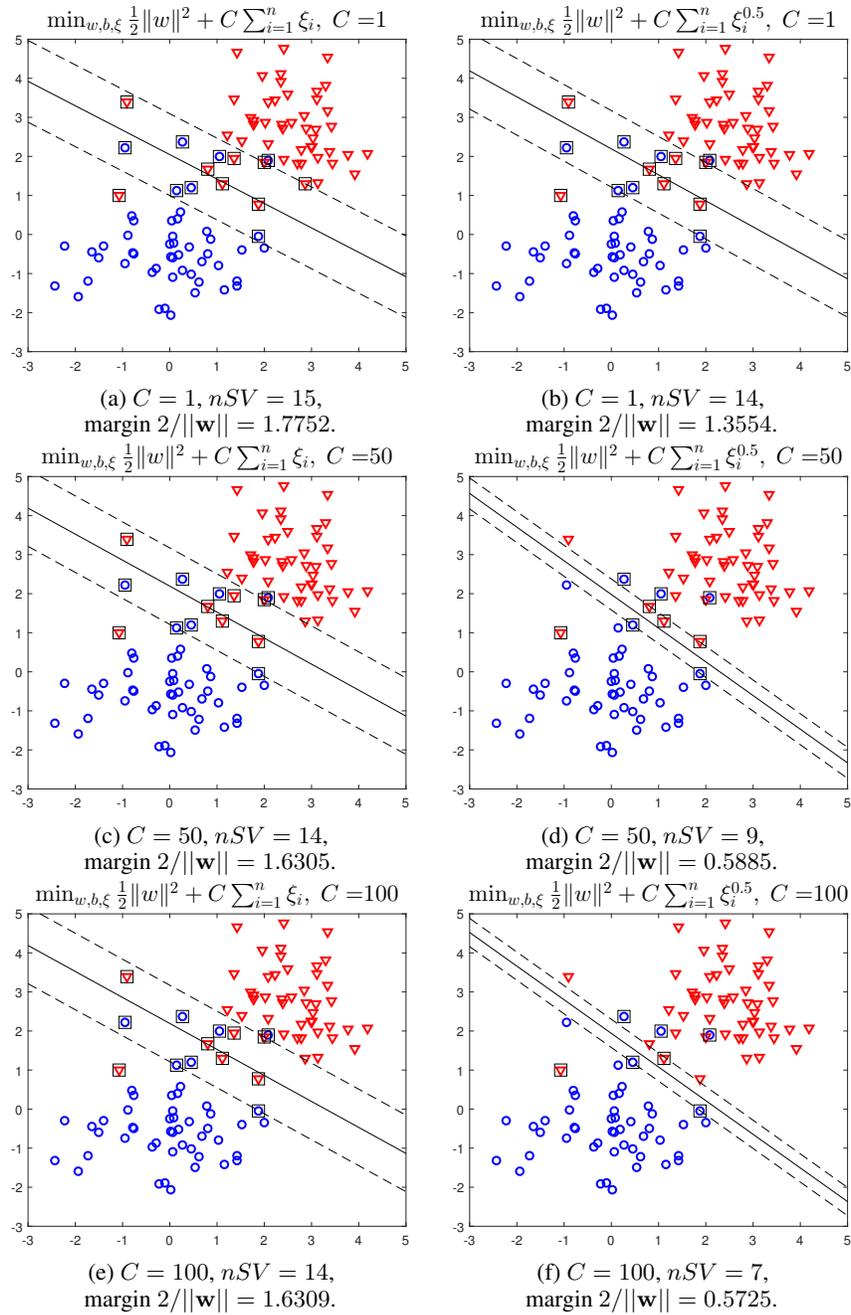

Figure 1: Comparison of SVM objective Eq.(8) and Eq.(9) on toy data ($nSV$ is number of support vectors).

When $p = 1$, Eq.(9) is the same as Eq.(8). When $p \to 0$, $\sum_i \xi_i^p$ approaches the number of nonzeros for $\xi_i, \forall i$. At small $p$, Eq.(9) will reduce number of nonzero $\xi_i$ and the number of support vectors.

The primal Lagrangian of Eq.(9) is:

$$L_P = \frac{1}{2}\|\mathbf{w}\|^2 + C\sum_i \xi_i^p - \sum_i \alpha_i\{y_i(\mathbf{w}^T\mathbf{x}_i + b) - 1 + \xi_i\} - \sum_i \mu_i\xi_i, \quad (10)$$

where $\alpha_i$ and $\xi_i$ are the Lagrange multipliers to enforce the positive constraints. The KKT conditions for the primal problem are given as:

$$\frac{\partial L_P}{\partial \mathbf{w}} = \mathbf{w} - \sum_i \alpha_i y_i \mathbf{x}_i = 0, \quad (11)$$

$$\frac{\partial L_P}{\partial b} = -\sum_i \alpha_i y_i = 0, \quad (12)$$

$$y_i(\mathbf{w}^T\mathbf{x}_i + b) - 1 + \xi_i \geq 0, \quad (13)$$

$$\xi_i \geq 0, \quad (14)$$
$$\alpha_i \geq 0, \quad (15)$$
$$\mu_i \geq 0, \quad (16)$$
$$\alpha_i\{y_i(\mathbf{w}^T\mathbf{x}_i + b) - 1 + \xi_i\} = 0, \quad (17)$$
$$\xi_i(pC\xi_i^{p-1} - \alpha_i) = 0. \quad (18)$$

$\mathbf{x}_i^T$ is the transpose of row vector $\mathbf{x}_i$. Eq.(17, 18) are KKT complementarity conditions. Eq.(17) is the same as Eq.(55) in [1]. We can get Eq.(18) using $\partial L_P/\partial \xi_i = 0$ and $\mu_i\xi_i = 0$.

For ease of notation, we append $b$ to vector $\mathbf{w}$ and append value 1 to $\mathbf{x}_i$

$$\mathbf{w}' = [\mathbf{w}, b] \quad (19)$$
$$\mathbf{x}'_i = [\mathbf{x}_i, 1] \quad (20)$$

Using Eqs.(14, 15, 17), Eq.(9) becomes a function with respect to vector $\mathbf{w}'$. When $\alpha_i > 0$, we have the following equation:

$$\xi_i = (1 - y_i\mathbf{w}'^T\mathbf{x}'_i)_+, \quad (21)$$

where, for a number $x$, when $x > 0$, $(x)_+ = x$; when $x <= 0$, $(x)_+ = 0$. When $\alpha_i = 0$, from Eq.(18), we have $pC\xi_i^p = 0$, which implies $\xi_i = 0$.

Using Eq.(21), Eq.(9) becomes:

$$\min \frac{1}{2}\mathbf{w}'^T D\mathbf{w}' + C\sum_i (1 - y_i\mathbf{w}'^T\mathbf{x}'_i)_+^p, \quad (22)$$

where $D \in \mathbb{R}^{(k+1)\times(k+1)}$ is an identity matrix with the last diagonal element $D(k+1, k+1)$ being 0. Eq.(22) can be solved using gradient descent with momentum [4].

**Algorithm** Since the derivative of function $(x)_+$ is not well defined when $x = 0$, we use the auxiliary function

$$(x)_+ = \lim_{s \to +\infty} \frac{1}{s}\log(1 + \exp sx), \quad (23)$$

where $s$ is a large number, for example, $s = 100, s = 200$.

The gradient of Eq.(22) is:

$$\nabla J(\mathbf{w}') = D\mathbf{w}' - pC\sum_i \frac{y_i m_i n_i^{p-1}}{1 + m_i}\mathbf{x}'_i, \quad (24)$$

where

$$m_i = \exp s(1 - y_i\mathbf{w}'^T\mathbf{x}'_i), \quad (25)$$
$$n_i = \frac{1}{s}\log(1 + m_i). \quad (26)$$

Algorithm 1: Gradient descent with Momentum to solve Eq.(9).

**Input:** Training data and label $\{\mathbf{x}_i, y_i\}$, $i = 1, ..., n$, parameter $C$, learning rate $\eta$, momentum coefficient $\varepsilon$
**Output:** $\mathbf{w}, b$
1: Initialize $\mathbf{w}, b, \mathbf{v}_0$
2: Form $\mathbf{w}'$ and $\mathbf{x}'_i$ using Eqs.(19, 20)
3: **while** Not converge **do**
4:     Compute gradient using Eq.(24)
5:     Compute $\mathbf{v}$ using Eq.(27)
6:     Update $\mathbf{w}'$ using Eq.(28)
7: **end while**

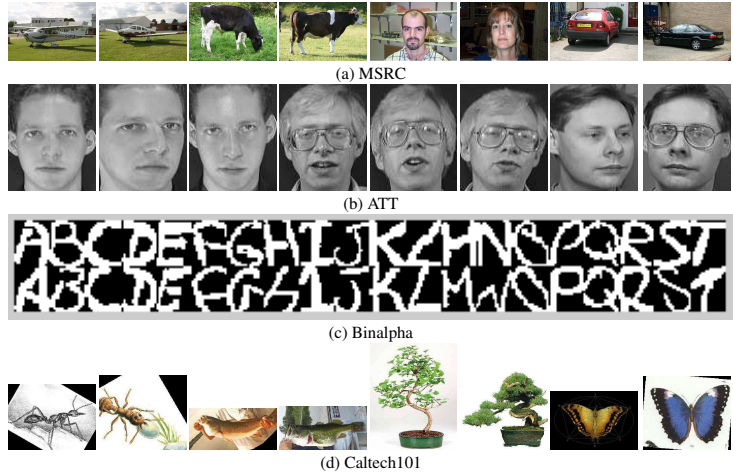

(a) MSRC

(b) ATT

(c) Binalpha

(d) Caltech101

Figure 2: Experiment example images.

Let $\eta > 0$ be the learning rate, $\varepsilon \in [0, 1]$ be the momentum coefficient, $\nabla J(\mathbf{w}'_t)$ be the gradient of Eq.(22) at iteration $t$.

$$\mathbf{v}_{t+1} = \varepsilon \mathbf{v}_t - \eta \nabla J(\mathbf{w}'_t), \tag{27}$$
$$\mathbf{w}'_{t+1} = \mathbf{w}'_t + \mathbf{v}_{t+1}, \tag{28}$$

$\mathbf{v}_t$ is initialized as vector of zeros. When optimal $\mathbf{w}'$ is found, we can get $\mathbf{w}$ and $b$ using Eq.(19).

Algorithm 1 summarizes the steps to solve Eq.(9). Using the solution $\mathbf{w}$ and $b$ of Algorithm 1, testing data $\mathbf{x}$ can be classified using $sign(\mathbf{w}^T \mathbf{x} + b)$. Support vectors are those points with positive $\xi_i$ computed from Eq.(21).

Figure 1b, 1d, 1f are the results of applying objective Eq.(9) with $p = 0.5$ on the same toy data. We can see that the number of support vectors is reduced significantly when $C$ increases from 1 to 100.

## 4 Experiments

In experiments, we select 7 binary classifcation tasks from 4 data sets as examples. We use $p = 0.5$ and study the convergence of Algorithm 1 and compare the classification performance of Minimal SVM and standard SVM.

### 4.1 Data

Four image datasets are used in this experiment. Data attributes are summarized in Table 1. Example images are shown in Figure 2.

**MSRC**[6] is an image scene data from MSRC data base v1, which includes tree, building, plane, cow, face, car and so on. 432-dimensional HOG feature is used in this chapter.

Table 1: Data attributes.

| Data | Dimension | Number of points in each class |
|---|---|---|
| MSRC | 432 | 30 |
| ATT | 644 | 10 |
| Binalpha | 320 | 39 |
| Caltech101 | 432 | 30 |

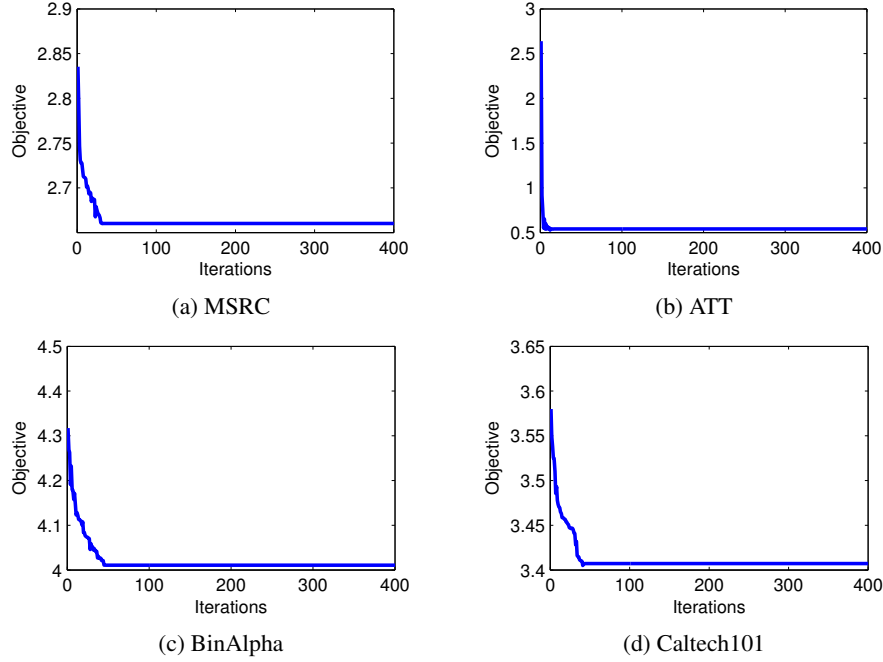

(a) MSRC

(b) ATT

(c) BinAlpha

(d) Caltech101

Figure 3: Objective function Eq.(9) converges using Algorithm 1.

**ATT** [3] data contains 400 images of 40 persons, with 10 images for each person. The images has been resized to $28 \times 23$ pixels.

**Binalpha** data contains 26 binary hand-written alphabets. We use the 320-dimensional pixels feature.

**Caltech101** [2] contains 101 object categories. We use the 432-dimensional HOG feature in this chapter.

### 4.2 Convergence of Algorithm

Algorithm 1 is very efficient on the experiment datasets. Figure 3 shows Algorithm 1 on the four datasets converges in less than 50 iterations.

### 4.3 Evaluation

Table 2 shows the evaluation results using four data sets. Each column is a two-class classification experiments using standard SVM Eq.(8) solution $\mathbf{w}_{L1}$ and Minimal SVM Eq.(9) solution $\mathbf{w}_{L05}$ with $p = 0.5$. We compare the classification accuracy of testing, training and number of support vectors (# SV). Angle $\theta$ measures the angle degree between $\mathbf{w}_{L1}$ and $\mathbf{w}_{L05}$:

$$\theta = \arccos \frac{\mathbf{w}_{L1} \cdot \mathbf{w}_{L05}}{||\mathbf{w}_{L1}||||\mathbf{w}_{L05}||} \frac{180}{\pi}. \tag{29}$$

Distance $d$ is the normalized Euclidean distance computed as:

$$d = \frac{||\mathbf{w}_{L1} - \mathbf{w}_{L05}||}{||\mathbf{w}_{L1}||}. \tag{30}$$

Table 2: Experiment results ($p = 0.5$).

|  |  | MSRC | ATT | BinAlpha | Caltech101 | | | |
|---|---|---|---|---|---|---|---|---|
| SVM | Test Acc | 0.67 | 0.85 | 0.89 | 0.70 | 0.90 | 0.57 | 0.73 |
|  | Train Acc | **0.95** | **1.00** | **0.99** | **0.99** | **1.00** | **0.98** | **1.00** |
|  | # SV | 38.20 | 11.00 | 53.00 | 42.20 | 32.40 | 43.40 | 36.40 |
| Minimal SVM | Test Acc | **0.72** | **0.90** | **0.90** | **0.73** | **0.92** | **0.58** | **0.75** |
|  | Train Acc | **0.95** | **1.00** | **0.99** | 0.97 | **1.00** | 0.95 | 0.98 |
|  | # SV | **22.40** | **2.00** | **33.40** | **31.80** | **18.40** | **30.80** | **17.80** |
| Angle $\theta$ | | 5.95 | 1.87 | 5.72 | 5.60 | 3.16 | 6.75 | 1.99 |
| Dist $d$ | | 0.13 | 0.06 | 0.12 | 0.11 | 0.09 | 0.14 | 0.04 |

All experiments are the average of 5-fold cross validation results. The test accuracy and train accuracy number the is between 0 and 1, the larger the better. The number of support vectors are the smaller the better. Best results are in bold in Table 2. We can see that, Minimal SVM gives the best test classification on these two classes classification test and has much less support vectors compared to standard SVM. To further investigate the difference of $\mathbf{w}_{L1}$ and $\mathbf{w}_{L05}$, we found that the angle degree difference is between $1.87$ to $6.75$ degrees. The normalized Euclidean distance is between $0.04$ and $0.14$. Even though many big data technologies including cloud computing, dimension reduction, accelerating algorithms have been proposed [5, 8, 10, 11, 12, 14, 15], in many cases, SVM generalization performance is still considered state-of-art approach for classification and regression applications [1, 9, 13].

## 5 Conclusion

In this work, we proposed a Minimal SVM, which uses $L_p$ norm on slack variables. We solve the objective using gradient descent with momentum by introducing a smoothing auxilary function. On 7 binary classification tasks, the proposed model further reduces the number of support vectors and increases the classification accuracy compared to standard SVM.